\documentclass[10pt,twocolumn,letterpaper]{article}

\usepackage{cvpr}
\usepackage{times}
\usepackage{epsfig}
\usepackage{graphicx}
\usepackage{amsmath}
\usepackage{amssymb}

\usepackage[lofdepth,lotdepth]{subfig}
\usepackage{algorithm}
\usepackage{multirow}
\usepackage{float}
\usepackage{rotating}
\usepackage{caption}
\usepackage{tabularx}

\usepackage[noend]{algpseudocode}

\usepackage[pagebackref=true,breaklinks=true,letterpaper=true,colorlinks,bookmarks=false]{hyperref}
\usepackage[normalsections,normalmargins,normalindent,normalleading,normaltitle,normalbib]{savetrees}
\cvprfinalcopy % *** Uncomment this line for the final submission

\def\cvprPaperID{1391} % *** Enter the CVPR Paper ID here

\ifcvprfinal\fi

\newcommand {\bsigma}{\boldsymbol{\sigma}} %Bold-face lower-case sigma
\newcommand {\bSigma}{\boldsymbol{\Sigma}} %Bold-face upper-case sigma
\newcommand {\bM}{\boldsymbol{M}}          %Bold-face M
\newcommand {\bx}{\boldsymbol{x}}          %Bold-face lower-case x
\newcommand {\bb}{\boldsymbol{b}}

\newcommand {\bB}{\boldsymbol{B}}
\newcommand {\bC}{\boldsymbol{C}}
\newcommand {\bU}{\boldsymbol{U}}
\newcommand {\bV}{\boldsymbol{V}}
\newcommand {\bz}{\boldsymbol{z}}

\newcommand {\bI}{\boldsymbol{I}}          %Bold-face upper-case I
\newcommand {\bT}{\boldsymbol{T}}          %Bold-face upper-case T
\newcommand {\ba}{\boldsymbol{a}}          %Bold-face lower-case a
\newcommand {\bc}{\boldsymbol{c}}          %Bold-face lower-case c
\newcommand {\by}{\boldsymbol{y}}          %Bold-face lower-case y
\newcommand {\Cbar}{\bar{C}}               %C bar

\newcommand {\bu}{\boldsymbol{u}}          %Bold-face lower-case u
\newcommand {\R}{\mathbb {R}}              %Blackboard Bold R - Real Numbers
\renewcommand{\algorithmicrequire}{\textbf{Input:}}
\renewcommand{\algorithmicensure}{\textbf{Output:}}
\algrenewcommand{\algorithmicrequire}{\textbf{Input:}}
\algrenewcommand{\algorithmicensure}{\textbf{Output:}}

\begin{document}

%%%%%%%%% TITLE
\title{Better Feature Tracking Through Subspace Constraints}

\author{Bryan Poling\\
University Of Minnesota\\
{\tt\small poli0048@umn.edu}
\and
Gilad Lerman\\
University Of Minnesota\\
{\tt\small lerman@umn.edu}
\and
Arthur Szlam\\
City College of New York\\
{\tt\small aszlam@ccny.cuny.edu}
}

\maketitle
%\thispagestyle{empty}

%%%%%%%%% ABSTRACT
\begin{abstract}
Feature tracking in video is a crucial task in computer vision.
%- used extensively in areas like navigation, SLAM, motion segmentation, and structure from motion.
Usually, the tracking problem is handled one feature at a time, using a single-feature tracker like the Kanade-Lucas-Tomasi algorithm, or one of its derivatives.
While this approach works quite well when dealing with high-quality video and ``strong'' features, it often falters when faced with dark and noisy video containing low-quality features. We present a framework for jointly tracking a set of features, which enables sharing information between the different features in the scene.
%Locally-rigid environment include several scenarios: single rigid body motion (where it can be simply justified), some non-rigid single motions (though locally rigid in small time frames) and motions of multiple rigid bodies or multiple bodies with locally rigid motion in time.
We show that our method can be employed to track features for both rigid and nonrigid motions (possibly of few moving bodies) even when some features are occluded. Furthermore, it can be used to significantly improve tracking results in poorly-lit scenes (where there is a mix of good and bad features). Our approach does not require direct modeling of the structure or the motion of the scene, and runs in real time on a single CPU core.
\end{abstract}

%%%%%%%%% BODY TEXT
\section{Introduction}

Feature tracking in video is an important computer vision task, often used as the first step in finding structure from motion or simultaneous location and mapping (SLAM). The celebrated Kanade-Lucas-Tomasi algorithm~\cite{lucas1981iterative, Tomasi91detectionand,shi_tomasi94} tracks feature points by searching for matches between templates representing each feature and a frame of video.
%The original Lucas-Kanade algorithm~\cite{lucas1981iterative} was suggested for estimating optical flow and was later adapted and applied to tracking~\cite{Tomasi91detectionand, black98, hager98}.
Despite many other alternatives and improvement, it is still one of the best video feature tracking algorithms~\cite{baker2004lucas}.\footnote{Feature tracking should be distinguished from object tracking, where there has been significant progress in the development of novel algorithms that significantly improve previous efforts.\\[0.25cm]
{\bf Acknowledgements:} This work was supported by NSF award DMS-09-56072, the Sloan foundation, the University of Minnesota Doctoral Dissertation Fellowship Program, and the Feinberg Foundation Visiting Faculty Program Fellowship of the Weizmann Institute of Science.\\[0.25cm]
{\bf Supp. web page:} http://www.math.umn.edu/$\sim$lerman/RCTracking/}
However, there are several realistic scenarios when Lucas-Kanade and many of its alternatives do not perform well: poor lighting conditions, noisy video, and when there are transient occlusions that need to be ignored. In order to deal with such scenarios more robustly it would be useful to allow the feature points to communicate with each other to decide how they should move as a group, so as to respect the underlying three dimensional geometry of the scene.

This underlying geometry constrains the trajectories of the track points to have a low-rank structure; see ~\cite{Costeira98, Hartley2000} for the case when tracking a single rigid object under an affine camera model, and ~\cite{BB01nonrigid, Torresani_nonrigid_flow2001, Hartley_Vidal_08, dai_cvpr2012} for non-rigid motion and the perspective camera.
%, but the rank of the cohort of trajectories may be higher.
In this work we will combine the low-rank geometry of the cohort of tracked features with the successful non-linear single feature tracking framework of Lucas and Kanade~\cite{lucas1981iterative} by adding a low-rank regularization penalty in the tracking optimization problem.
 To accommodate dynamic scenes with non-trivial motion we apply our rank constraint over a sliding window, so that we only consider a small number of frames at a given time (this is a common idea for dealing with non-rigid motions~\cite{Buchanan_Fitzgibbon_2007, Ricco_Tomasi_CVPR12, Garg_weak_constraint2013}).
We demonstrate very strong performance in rigid environments as well as in scenes with multiple and/or non-rigid motion (since the trajectories of all features are still low rank for short time intervals).
We describe experiments with several choices of low-rank regularizers (which are local in time), using a unified optimization framework that allows real time regularized tracking on a single CPU core.
%, and show that the recent notion of empirical dimension~\cite{gdm13} is especially suited for this task.

\subsection{Relationship With Previous Work}
%enforce the underlying geometry of the feature points of the tracked objects.
Geometric structures (and low-rank structures in particular) have been effectively utilized for the problem of optical flow estimation.
Irani~\cite{irani02} showed how 3-d constraints in the real world and various camera models imply the existence of a low-rank constraint on the flow problem.  Brand extended Irani's work to non-rigid motions, while developing a robust, subspace-estimating, flow-based tracker via an incremental singular value decomposition with missing values as well as by learning an object model~\cite{Brand02subspacemappings, Brand02incremental, Brand01morph, BB01nonrigid}. Torresani et al.~\cite{Torresani_nonrigid_flow2001} also extended Irani's work to non-rigid motions by applying rank-bounds for recovering 3D non-rigid motions.
More recently, Garg et al.~\cite{Garg_strong_constraint2010} introduced hard subspace constraints for long range optical flow estimation in a variational scheme.
Garg et al.~\cite{Garg_weak_constraint2013} improved the performance of this former work by making the constraint weak (as an energy regularizer), using a robust energy term and allowing more general basis terms. At last, Ricco and Tomasi~\cite{Ricco_Tomasi_CVPR12} proposed a Lagrangian approach for long range motion estimation that allows more reliable detection of occlusion. It estimates a basis for a low-dimensional subspace of the trajectories (as in~\cite{Garg_weak_constraint2013}) and employs a variational method to solve for the best-fit coefficients of the motion trajectories in this basis.
%At last, Rubinstein et al.~\cite{Rubinstein_low_range12}
%proposed a novel divide and conquer approach to long-range motion estimation
%with dense feature trajectories, which is robust to occlusion, deformation and camera motion.

In optical flow estimation the goal is to find displacements of features between consecutive frames, while assuming that the flow field is locally nearly constant.
%Irani showed that, given this formulation, a low rank projection of this system can be solved to enforce rigidity constraints.
Although the goal in the feature tracking problem is similar, it does not require estimating the flow by enforcing the brightness constancy constraint or a weaker version of it.  The subspace constraints above were translated by Irani~\cite{irani02} to an image brightness constraint.  However, small errors in the flow field in each frame from this approach lead to the accumulation of errors in the trajectories obtained by integrating the flow.  These errors are  unacceptable for tracking. Weak versions of this constraint for estimating flow along many frames (as in~\cite{BB01nonrigid, Ricco_Tomasi_CVPR12, Garg_weak_constraint2013}) require rather dense trajectories, which represent continuous regions in the image frame. Indeed, they are based on either continuous variational methods~\cite{Ricco_Tomasi_CVPR12, Garg_weak_constraint2013} (which often track all pixels in the image domain) or careful model estimation~\cite{BB01nonrigid} (which requires sufficiently dense sampling from objects in the videos).

In tracking, one instead uses a formulation that allows for very precise feature registration (like the Lucas-Kanade tracker~\cite{lucas1981iterative}), and there is no need to linearize the image to solve an approximation to the feature displacement problem. It is desirable to have a sparse set of features and track them only in local neighborhoods to allow real time implementation.
%The feature tracking formulation is significantly different from what is used for optical flow, and in particular, .
There is not a canonical method for introducing an explicit low rank constraint as in~\cite{irani02}. We will argue below that any strict subspace constraint is not ideal in the tracking problem and will promote a soft constraint. This soft constraint is different than the ones advocated for flow estimation in both~\cite{Ricco_Tomasi_CVPR12} and~\cite{Garg_weak_constraint2013} since they carefully learn local basis elements and require dense feature sampling.

Torresani and Bregler~\cite{Space-Time_Tracking2002} suggested the partial application of hard low-rank constraints to improve tracking (applying rank bounds as in~\cite{Torresani_nonrigid_flow2001}). They rely on initial Lucas-Kanade tracking~\cite{lucas1981iterative} from which ``reliable'' features are identified and used to estimate a model for the scene. They used their constraint to re-track the ``unreliable'' features (the trajectories are now confined to a known subspace). Since they search in the space of trajectories, their minimization strategy is completely different than ours. Their tracker is also non-casual since it needs the full sequence to start tracking, so a real-time implementation is not possible. This work was extended in \cite{torresani2003robust} to develop a causal tracker in the same spirit that also does not rely on a set of ``reliable'' feature tracks. However, both methods require setting the rank of the constraint a-priori and they impose the constraint over very long time spans (up to the entire sequence), making the algorithms less applicable to dynamic scenes.

Buchanan and Fitzgibbon~\cite{Buchanan_Fitzgibbon_2007}
continuously update a non-rigid motion model over a sliding temporal
window. This motion model is
used as a motion prior in a conventional Bayesian template
tracker for a single feature.
The local information is combined with weaker
global low rank approximation for the set of initial local trajectories (in the spirit of~\cite{BB01nonrigid, Ricco_Tomasi_CVPR12, Garg_weak_constraint2013}, while different than the low rank constraint of this paper). Similarly to~\cite{BB01nonrigid} this low rank constraint
guides the tracking via Bayesian modeling.% (though applied here for possibly sparse features over long sequences).

%We are unaware of using the linear constraint for tracking of videos with many frames (and not for flow estimation or tracking with only few frames).
Another line of work takes tracked feature points in videos, and then uses the underlying subspace structure of rigid bodies to segment different motions of such bodies~\cite{Costeira98, Yan06ageneral, spectral_applied, LBF_journal12, ssc_elhamifar13}. This is related to the large body of work on recovering rigid or non-rigid structure from motion; see \cite{Hartley2000} or \cite{dai_cvpr2012} and the references therein.
However, these works are highly dependent on good tracking and it would be desirable to simultaneously track and segment motion, or exploit the subspace structure to improve tracking prior to finding structure from motion.
%; this work is a first step.

\section{On Low-Rank Feature Trajectories}
Under the affine camera model, the feature trajectories for a set of features
from a rigid body should exist in an affine subspace of dimension $3$, or
a linear subspace of dimension 4~\cite{Costeira98, Hartley2000}.
However, subspaces corresponding to very degenerate motion are lower-dimensional than those corresponding to general motion~\cite{Hartley2000}.

Feature trajectories of non-rigid scenarios exhibit significant variety, but some low-rank models may still be successfully applied to them~\cite{BB01nonrigid, Torresani_nonrigid_flow2001, Hartley_Vidal_08, dai_cvpr2012, Garg_weak_constraint2013}. Similarly to~\cite{Buchanan_Fitzgibbon_2007, Garg_weak_constraint2013} (though in a different setting) we consider a sliding temporal window, where over short durations the motion is simple and the feature trajectories are of lower rank.
The restriction on the length of feature trajectories can also help in satisfying an approximate local affine camera model in scenes which violate the affine camera model. In general, depth disparities give rise to low-dimensional manifolds~\cite{Hartley2000} which are only locally approximated by linear spaces.

At last, even in the case of multiple moving rigid objects, the set of trajectories is still low rank (confined to the union of a few low rank subspaces). %gives rise to low rank model (possibly local) as the union of the low rank models is low rank.
In all of these scenarios the low rank is unknown in general.

\section{Feature Tracking}
\label{sec:Tracking Features on a Rigid Body}

%\subsection*{Notation}
%We view an image (or frame of a video) as a function from $\R^2$ to $\R^D$, with $D=1$ or $D=3$, and thus may apply continuous location.
%However, we also express its voxel values at discretized locations.
{\bf Notation:} A \emph{feature} at a location $\bz_1 \in \R^2$ in a given $N_1 \times N_2$ frame of an $N_1 \times N_2 \times N_3$ video
is characterized by a \emph{template} $T$, which is an $n \times n$ sub-image of that frame centered at $\bz_1$ ($n$ is a small integer, generally taken to be odd, so the template has a center pixel). If $\bz_1$ does not have integer coordinates, $T$ is interpolated from the image.
We denote $\Omega=\{1,...,n\}\times\{1,...,n\}$ and we parametrize $T$ so that its pixel values are obtained by $\{T(\bu)\}_{\bu \in \Omega}$. %(with this parametrization $\bz_1$ will not be considered anymore in our notation).

%\subsection*{Tracking Formulation}

A classical formulation of the single-feature tracking problem (see e.g., \cite{lucas1981iterative}) is to search for the translation $\bx_1$ that minimizes some distance between a feature's template $T$ at a given frame and the next frame of video translated by $\bx_1$; we denote this next frame by $I$. That is, we minimize the \emph{single-feature energy function} $c(\bx_1)$:
\begin{equation}
\label{eqn:Single Feature Energy}
c(\bx_1) = {1 \over n^2} \sum_{\bu \in \Omega} \psi \left( T(\bu) - I(\bu + \bx_1) \right),
\end{equation}
where, for example, $\psi(x) = |x|$ or $\psi(x) = x^2$.
%... continuous minimization
To apply continuous optimization we view $\bx_1$ as a continuous variable and we thus view $T$ and $I$ as functions over continuous domains (implemented with bi-linear interpolation).

\subsection{Low Rank Regularization Framework}
\label{sec:Low rank regularization}

If we want to encourage a low rank structure in the trajectories, we cannot view the tracking of different features as separate problems.
%To this end, we combine the single-feature energy functions defined in \eqref{eqn:Single Feature Energy} to form the \emph{total energy} function.
For $f \in \{1,2,...,F\}$, let $\bx_f$ denote the position of feature $f$ in the current frame (in image coordinates), and let $\bx = (\bx_1,\bx_2,...,\bx_F) \in \R^{2F}$ denote the joint state of all features in the scene. We define the total energy function as follows:
\begin{equation}
\label{eqn:Total Energy Without Constraint}
C(\bx) = {1 \over Fn^2} \sum_{f=1}^{F} \sum_{\bu \in \Omega} \psi \left( T_f(\bu) - I(\bu +\bx_f) \right),
\end{equation}
where $T_f(\bu)$ is the template for feature $f$.
Now, we can impose desired relationships between features in a scene by imposing constraints on the domain of optimization of \eqref{eqn:Total Energy Without Constraint}. %Next we discuss why a low-rank constraint is natural for various scenarios and propose some strategies for imposing it.

%\section{Introducing Low-Rank Constraints to Features' Tracking}

Instead of enforcing a hard constraint, we add a \emph{penalty term} to \eqref{eqn:Total Energy Without Constraint}, which increases the cost of states which are inconsistent with low-rank motion. Specifically, we define:
\begin{equation}
\label{eqn:Total Energy}
\bar{C}(\bx) = \alpha \sum_{f=1}^{F} \sum_{\bu \in \Omega} \psi \left( T_f(\bu) - I(\bu + \bx_f) \right) + P(\bx),
\end{equation}
where $P(\bx) $ is an estimate of, or proxy for, the dimensionality of the set of feature trajectories over the last several frames of video (past feature locations are treated as constants, so this is a function only of the current state, $\bx$).
Notice that we have replaced the scale factor $1/(Fn^2)$ from \eqref{eqn:Total Energy Without Constraint} with the constant $\alpha$, as this coefficient is now also responsible for controlling the relative strength of the penalty term.
We will give explicit examples for $P$ in section \ref{sec:Specific Choices of the Low-Rank Regularizer}.

This framework gives rise to two different solutions, characterized by the strength of the penalty term (definition of $\alpha$). Each has useful, real-world tracking applications.
In the first case, we assume that most (but not necessarily all) features in the scene approximately obey a low rank model. This is appropriate if the scene contains non-rigid or multiple moving bodies. We can impose a \emph{weak constraint} by making the penalty term small relative to the other terms. If a feature is strong, it will confidently track the imagery, ignoring the constraint (regardless of whether the motion is consistent with the other features in the scene). If a feature is weak in the sense that we cannot fully determine its true location by only looking at the imagery, then the penalty term will become significant and encourage the feature to agree with the motion of the other features in the scene.

In the second case, we assume that all features in the scene are supposed to agree with a low rank model (and deviations from that model are indicative of tracking errors). We can impose a strong constraint by making the penalty term large relative to the other terms. No small set of features can overpower the constraint, regardless of how strong the features are. This forces all features to move is a way that is consistent with a simple motion. Thus, a small number of features can even be occluded, and their positions will be predicted by the motion of the other features in the scene. We further explain these two scenarios and demonstrate them with figures in the supplementary material.

\subsection{Specific Choices of the Low-Rank Regularizer}
\label{sec:Specific Choices of the Low-Rank Regularizer}
There is now a large body of work on low rank regularization, e.g., \cite{Candes_Recht09, wright_robust_pca09, Negahban12}.  We will restrict ourselves to showing results using three choices for $P$ described below. Each choice we present defines $P(\bx)$ in terms of a matrix
$\bM$. It is the $2(L+1) \times F$ matrix whose column $f$ contains the feature trajectory for feature $f$ within a sliding window of $L+1$ consecutive frames (current frame and $L$ past frames).
Specifically,
$\bM = \left[m_{i,j}\right]$, where $(m_{0,f}, m_{1,f})^T$ is the current (variable) position of feature $f$ and $(m_{2l+1,f}, m_{2l+2,f})^T$, $l=1,...,L$
contains the $x$ and $y$ pixel coordinates of feature $f$ from $l$ frames in the past (past feature locations are treated as known constants).
One may alternatively center the columns of $\bM$ by subtracting from each column the average of all columns. Most constraints derived for trajectories (assuming, for instance, rigid motion) actually confine trajectories to a low rank {\em affine} subspace (as opposed to a {\em linear} subspace). Centering the columns of $\bM$ transforms an affine constraint into a linear one. Alternatively, one can forgo centering and view an affine constraint as a linear constraint in one dimension higher. We report results for both approaches.
%{\em Centering} the columns of $\bM$ results in a constraint that trajectories live in a low rank {\em linear} subspace. If $\bM$ is not centered, trajectories are confined to a low rank {\em affine} subspace. We report results for both approaches.
%even though they are hardly any different in practice.
% (since the dimension is not known and thus not imposed on the problem, we indeed expect no significant difference).

\subsubsection*{Explicit Factorizations}
A simple method for enforcing the structure constraint is to  write ${\bM}=\bB \bC$, where $\bB$ is a $2(L+1)\times d$ matrix, and $\bC$ is a $d\times F$ matrix.  However, as mentioned in the previous section, because the feature tracks often do not lie exactly on a subspace due to deviations from the camera model or non-rigidity, an explicit constraint of this form is not suitable.

However, an explicit factorization can be used in a penalty term by measuring the deviation of ${\bM}$, in some norm, from its approximate low rank factorization. For example, if we let \begin{equation}
{\bM}=\bU\bSigma \bV^T
\end{equation} denote the SVD of ${\bM}$, we can take $P(\bx)$ in  \eqref{eqn:Total Energy} to be
 $||\bB \bC-\bM||_*$, where $\bB$ is the first three or four columns of $\bU$, and $\bC$ is the first three or four rows of $\bSigma \bV^T$.  
%$\Pi(\bx)=\bB \balpha$, and $\balpha=\text{argmin}_{\bz} ||\bB \bz-\bx||^2$.
Then this $P$ corresponds  to penalizing ${\bM}$ via $\sum_{i=d+1}^F\sigma_i$, where $\sigma_i=\bSigma_{ii}$ is the $i$'th singular value of ${\bM}$.  As above, since the history is fixed, $\bU$, $\bSigma$, and $\bV^T$ are functions of $\bx$.%; in the experiments below we will alternate updates of the optimal $\bx$ and SVD of $\bM$ rather than exaustively compute the SVD for all possible $\bx$.

This approach is the closest analogue of \cite{irani02} in the tracking setting, next to an explicit rank constraint.
It assumes knowledge of the low-rank $d$. For simplicity, we assume a local rigid model and thus set $d=3$ when centering $M$ and $d=4$ when not centering (following~\cite{Costeira98, Hartley2000}).
%While we advocate for other methods, where the low rank is unknown, we use it for the purpose of comparison and understanding the difference between the different methods.  Why do we advocate for other methods- if we say this, need to justify...
\subsubsection*{Nuclear Norm}
A popular alternative to explicitly keeping track of the best fit low-dimensional subspace to ${\bM}$ is to use the matrix nuclear norm and define
\begin{equation}
P(\bx) = \Vert {\bM} \Vert_* = \Vert \bsigma \Vert_1.
\end{equation}
This is a convex proxy for the rank of ${\bM}$ (see e.g., \cite{Candes_Recht09, wright_robust_pca09}).  Here $\bsigma = (\sigma_1 \; \sigma_2 \; \ldots \; \sigma_{2(L+1) \land F})^T$  is the vector of singular values of ${\bM}$, and $||\cdot ||_1$ is the $l_1$ norm. Unlike explicit factorization, where only energy outside the first $d$ principal components of ${\bM}$ is punished,
the nuclear norm will favor lower-rank ${\bM}$ over higher-rank ${\bM}$ even when both matrices have rank $\le d$. %, even when considering matrices of rank $\le 3$.
Thus, using this kind of penalty will favor simpler track point motions over more complex ones, even when both are technically permissible.

\subsubsection*{Empirical Dimension}
\label{sec:Empirical Dimension}
Empirical Dimension~\cite{gdm13} refers to a class of dimension estimators depending on a parameter $\epsilon \in (0,1]$.
%Let $\A$ be a $D \times N$ data matrix where each column is a data vector. We denote by $\bsigma = (\sigma_1 \; \sigma_2 \; \ldots \; \sigma_{N \land D})^T$  the vector of singular values of $\A$. Then,
The empirical dimension of ${\bM}$ is defined to be:
\begin{equation}
\label{eqn:Empirical Dimension Definition}
\hat{d}_\epsilon({\bM}) := {{\Vert \bsigma \Vert_\epsilon} \over {\Vert \bsigma \Vert_{\left({{\epsilon} \over {1-\epsilon}}\right)}}}.
\end{equation}
Notice that we use norm notation, although $\Vert \cdot \Vert_\epsilon$ is only a pseudo-norm. When $\epsilon=1$, this is sometimes called the ``effective rank'' of the data matrix~\cite{vershynin_book}.

Empirical dimension satisfies a few important properties, which are verified in~\cite{gdm13}. First, empirical dimension is invariant under rotation and scaling of a data set. Additionally, in the absence of noise, empirical dimension never exceeds true dimension, but it approaches true dimension as the number of measurements goes to infinity for spherically symmetric distributions. Thus, $d_\epsilon$ is a true dimension estimator (whereas the nuclear norm is a proxy for dimension). To use empirical dimension as our regularizer, we define $P(\bx) = d_\epsilon({\bM})$.

Empirical dimension is governed by its parameter, $\epsilon$. An $\epsilon$ near $0$ results in a ``strict'' estimator, which is appropriate for estimating dimension in situations where you have little noise and you expect your data to live in true linear spaces. If $\epsilon$ is near $1$ then $d_\epsilon$ is a lenient estimator. This makes it less sensitive to noise, and more tolerant of data sets that are only approximately linear. In all of the experiments we present, we use $\epsilon = 0.6$, although we found that other tested values also worked well.
%A thorough development of empirical dimension (including proofs of the properties mentioned above) is provided in~\cite{gdm13}.

%For any value of $\epsilon$ in $(0,1]$, empirical dimension is an almost-everywhere differentiable function of the elements of $\A$. This gives it potential for variational optimization applications, and it will be essential in our tracking application.

\subsection{Implementation Details}
\label{sec:Implementation Details}
%In this section we address the implementation of a strong or weak constraint, and we describe the details of the gradient descent method we use to minimize the objective function \eqref{eqn:Total Energy}.
We fix $L=10$ for the sliding window and let $\psi(x) = |x|$ in \eqref{eqn:Total Energy}.
%With this scaling, $\psi$ is convex and the energy function is independent of the intensity range used to represent our imagery.
We use this form for $\psi$ so that all terms in the total energy function behave linearly in a known range of values. If our fit terms behaved quadratically, it would be more challenging to balance them against a penalty term. We also tested a Huber loss function for $\psi$ and have concluded that such a regularization is not needed.

We fix a parameter $m$ for each penalty form (selected empirically - see the supplementary material for our procedure), which determines the strength of the penalty.
The weak and strong regularization parameters are set as follows:
\begin{equation}
\label{eqn:alpha}
\alpha_{weak} = {1 \over m n^2}
\ \text{ and } \
\alpha_{strong} = {1 \over mFn^2}.
\end{equation}
The weak scaling implies that a perfectly-matched feature will contribute $0$ to the total energy, and a poorly-matched feature will contribute an amount on the order of $1/m$ to the total energy. The penalty term will contribute on the order of $1$ to the total energy.
Since we do not divide the contributions of each feature by the number of features, the penalty terms contribution is comparable in magnitude to that of a single feature.
The strong scaling implies that the penalty term is on the same scale as the sum of the contributions of all of the features in the scene.

\subsection*{Minimization Strategy}
The total energy function we propose for constrained tracking is non-convex  since the contributions from the template fit terms are not convex (even if $P$ is convex); this is also the case with other feature tracking methods, including the Lucas-Kanade tracker. We employ a $1^{\text{st}}$-order descent approach for driving the energy to a local minimum.

To reduce the computational load of feature tracking, some trackers use $2^{\text{nd}}$-order methods for optimization (see \cite{baker2004lucas}). This works well when tracking strong features, but in our experience it can be unreliable when dealing with weak or ambiguous features. Since we are explicitly trying to improve tracking accuracy on poor features we opt for a $1^{\text{st}}$-order descent approach instead.

The simplest $1^{\text{st}}$-order descent method is (sub)gradient descent. Unfortunately, because there can be a very large difference in magnitude between the contributions of strong and weak features to our total energy, our problem is not well-conditioned. If we pursue standard gradient descent, the strong features dictate the step direction and the weak features have very little effect on it. Ideally, once the strong features are correctly positioned, they will no longer dominate the step direction. If we were able to perfectly measure the gradient of our objective function, this would be the case. In practice, the error in our numerical gradient estimate can be large enough to prevent the strong features from ever relinquishing control over the step direction. The result is that in a scene with both very strong and very weak features, the weak features may not be tracked.

To remedy this, we compute our step direction by blending the gradient of the energy function with a vector that corresponds to taking equal-sized gradient descent steps separately for each feature. We use a fast line search in each iteration to find the nearest local minimum in the step direction. This compromise approach allows for efficient descent while ensuring that each feature has some control over the step direction (regardless of feature strength).

Because the energy is  not convex, it is important to choose a good initial state. We use a combination of two strategies to initialize the tracking: first, we generate our initial guess of $\bx$ by registering an entire frame of video with the previous (at lower resolution). Secondly, we use multi-resolution, or pyramidal tracking so that approximate motion on a large scale can help us get close to the minimum before we try tracking on finer resolution levels (see~\cite{Bergen92pyramid}).

We now explain the details of the algorithm. Let $\bI$ denote a full new frame of video and let $\bx^{\text{prev}}$ be the concatenation of feature positions in the previous frame. We form a pyramid for $\bI$ where level $0$ is the full-resolution image and each higher level $m$ ($1$ through $3$) has half the vertical and half the horizontal resolution of level $m-1$. To initialize the optimization, we take the full frame (at resolution level $3$) and register it against the previous frame (also at resolution level $3$) using gradient descent and an absolute value loss function. We initialize each features position in the current frame by taking its position in the previous frame and adding the offset between the frames, as found through this registration process). Once we have our initial $\bx$, we begin optimization on the top pyramid level. When done on the top level, we use the result to initialize optimization on the level below it, and so on until we have found a local minimum on level $0$. On any given pyramid level, we perform optimization by iteratively computing a step direction and conducting a fast line search to find a local minimum in the search direction. We impose a minimum and maximum on the number of steps to be performed on each level ($\text{min}_i$ and $\text{max}_i$, respectively). Our termination condition (on a given level) is when the magnitude of the derivative of $\Cbar$ is not significantly smaller than it was in the previous step. To compute our search direction in each step, we first compute the gradient of $\Cbar$ (which we will call $D\Cbar$) and set $\ba = -D\Cbar$. We then compute a semi-normalized version of $\ba$. This is done by breaking it into a collection of $2$-vectors (elements $1$ and $2$ are together, elements $3$ and $4$ are together, and so on) and normalizing each of them. We then re-combine the normalized $2$-vectors to get $\bb$. We blend $\ba$ with $\bc$ to compute our step direction. Algorithm \ref{Algorithm} summarizes the full process. Source code for our implementation of this algorithm will be available on the first authors web page.

%
%Algorithm table
\begin{algorithm}[ht]
\caption{Optimization of rank-penalized energy}
\begin{algorithmic}
\Require $\bx^{\text{prev}}$, $\bI$, $\bT_f \; f \in \{1,2,...,F \}$, $\bM$, $\alpha$, $\text{min}_i$, $\text{max}_i$
\Ensure  $\bx$
	\item Initialize $\bx = \bx^{\text{prev}} + [\Delta \bx, \Delta \bx, ..., \Delta \bx]^T$ where $\Delta \bx$ is the result of registering $\bI$ against the previous frame. 
	\For {$m=3:0$}
		\State $\bx \leftarrow (1/2)^m \bx$
		\State $\Vert D\Cbar \Vert_{\text{old}} \leftarrow \infty$
		\For {$i=1:\text{max}_i$}		
			\State Let $\ba \leftarrow -D\Cbar$
			\For {$f=1:F$}
				\State $\by_f \leftarrow [\ba(2f-1),\ba(2f)]^T$
				\State $\bb_f \leftarrow \by_f/|\by_f|$
			\EndFor
			\State $\bb \leftarrow [\bb_1^T,\bb_2^T,...,\bb_F^T]^T$
			\State $\bc \leftarrow 0.5\ba + 0.5\bb$
			\State $\bx = \bx + \bc d$ where $d$ is output of line search
			\If {$\Vert D\Cbar \Vert > 0.99 \Vert D\Cbar \Vert_{\text{old}}$ and $i>\text{min}_i$}
				\State Exit for loop early
			\Else
				\State Assign $\Vert D\Cbar \Vert_{\text{old}} = \Vert D\Cbar \Vert$
			\EndIf
		\EndFor
		\State $\bx \leftarrow 2^m \bx$
	\EndFor
\State Return $\bx$
\end{algorithmic}
\label{Algorithm}
\end{algorithm}

\subsection*{Efficiency and Complexity}
We have found that our algorithm typically converges in about 20 iterations or less at each pyramid level (with fewer iterations on lower pyramid levels). In our experiments, we used a resolution of $640$-by-$480$ (we have also done tests at $1000 \times 562$), and we found that 4 pyramid levels were sufficient for reliable tracking. Thus, on average, less than 80 iterations are required to track from one frame to the next. A single iteration requires one gradient evaluation and multiple evaluations of $\Cbar$. The complexity of a gradient evaluation is $k_1Fn^2 + k_2 L F^2$, and the complexity of an energy evaluation is $k_3 Fn^2 + k_4 L^2 F$ (details are given in the supplementary material). Our C++ implementation (which makes use of OpenCV) can run on $35$ features of size $7$-by-$7$ with a temporal window of 6 frames ($L=5$)\footnote{Accuracy for $L=5$ is only slightly worse than for $L=10$ and enables faster processing. See the supp. material for a brief comparison.} on a $3^{\text{rd}}$-generation Intel i5 CPU at approximately 16 frames per second. SIMD instructions are used in places, but no multi-threading was used, so faster processing rates are possible. With a larger window of $L=10$ our algorithm still runs at $2$-$5$ frames per second.

\section{Experiments}
\label{sec:Experiments}
To evaluate our method, we conducted tests on several real video sequences in circumstances that are difficult for feature tracking. These included shaky footage in low-light environments. The resulting videos contained dark regions with few good features and the unsteady camera motion and poor lighting introduced time-varying motion blur.

In these video sequences it proved very difficult to hand-register features for ground-truth. In order to present a quantitative numerical comparison we also collected higher-quality video sequences and synthetically degraded their quality. We used a standard Lucas-Kanade tracker on the non-degraded videos to generate ground-truth (the output was human-verified and corrected). We therefore present qualitative results on real, low-quality video sequences, as well as quantitative results on a set of synthetically degraded videos.

\begin{figure}[htbp]
\captionsetup{margin=10pt,labelfont=bf}
\centering
\subfloat[Dark Scene on frame 1]{
	\includegraphics[width=0.49\linewidth, clip=true, trim=0mm 0mm 0mm 0mm]{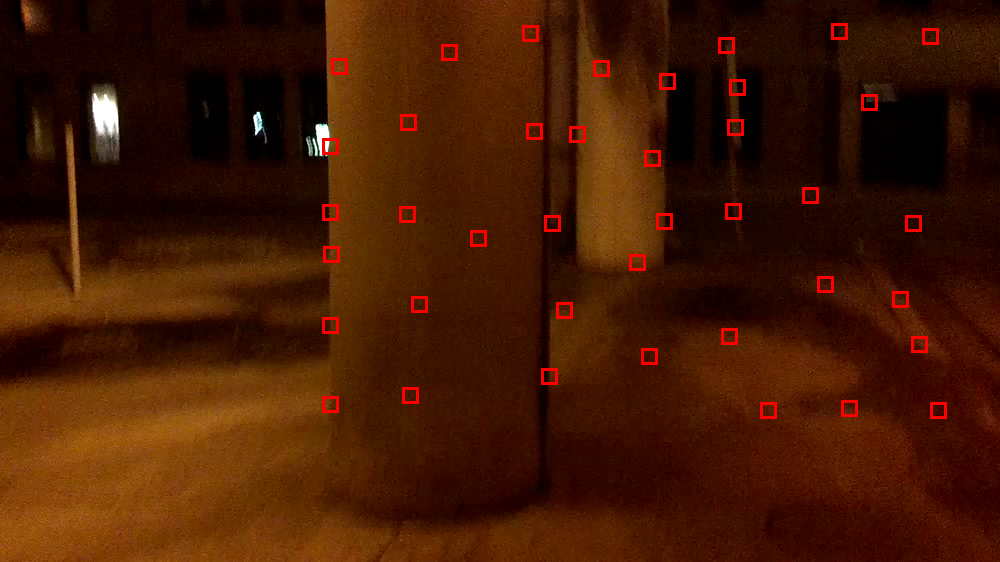}
}
\subfloat[Lucas Kanade - frame 30]{
	\includegraphics[width=0.49\linewidth, clip=true, trim=0mm 0mm 0mm 0mm]{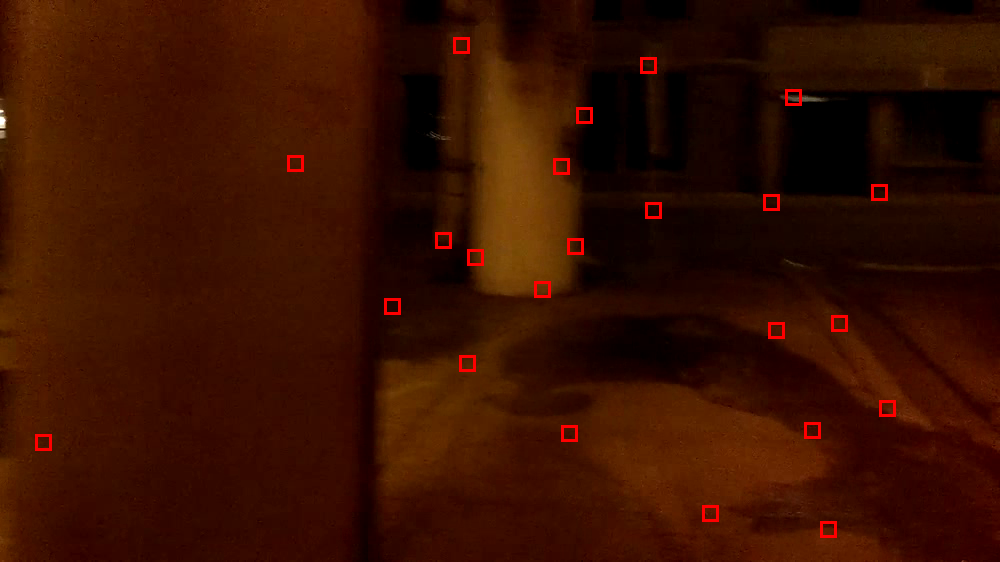}
}\\
\subfloat[Dark Scene on frame 1]{
	\includegraphics[width=0.49\linewidth, clip=true, trim=0mm 0mm 0mm 0mm]{Real-Frame1.png}
}
\subfloat[Our method - frame 30]{
	\includegraphics[width=0.49\linewidth, clip=true, trim=0mm 0mm 0mm 0mm]{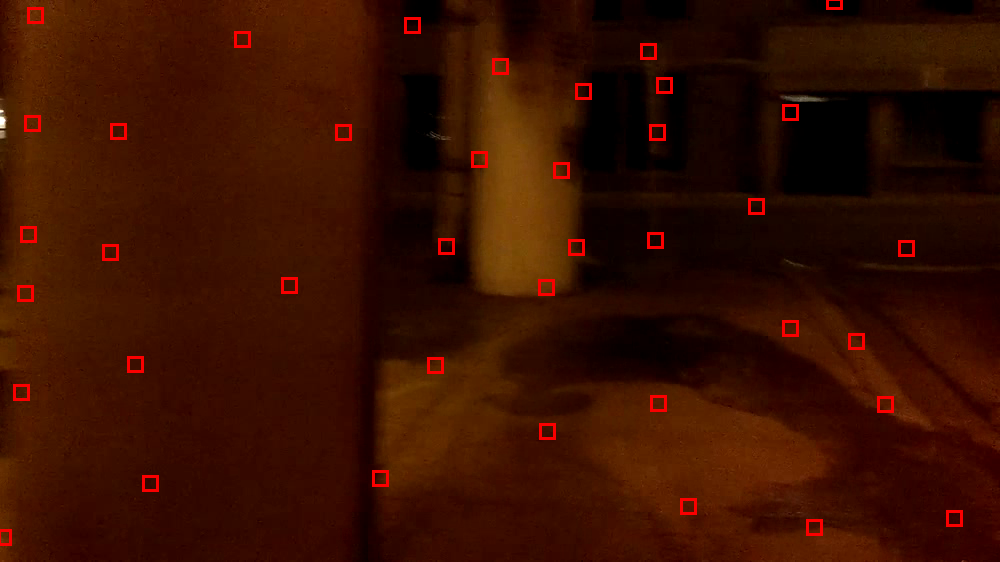}
}
\caption{\label{fig:Real Video}Results of tracking features in real low-light video. Most of the features wander significantly with the Lucas Kanade tracker. Our method provides better results on the low-quality features.}
\captionsetup{margin=10pt,labelfont=bf}
\centering
\subfloat[Lucas Kanade after 10 frames]{
	\includegraphics[width=\linewidth, clip=true, trim=32mm 24mm 32mm 24mm]{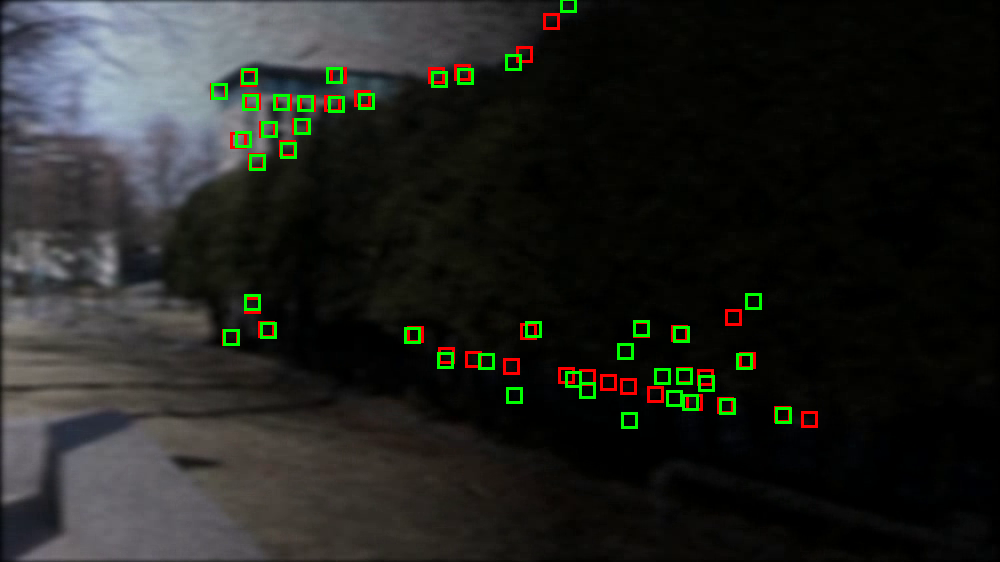}
}\\
\subfloat[Our method after 10 frames]{
	\includegraphics[width=\linewidth, clip=true, trim=32mm 24mm 32mm 24mm]{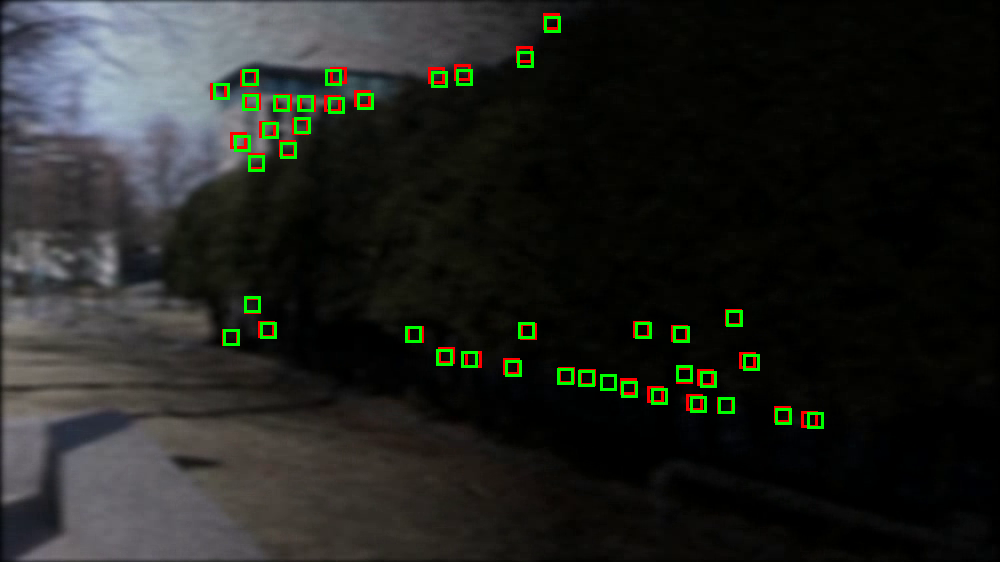}
}
\caption{\label{fig:Synthetic Experiment}Characteristic results of the OpenCV Lucas-Kanade tracker vs our method in our synthetically degraded video experiment. The correct feature locations (according to the Lucas-Kanade tracker on the non-degraded video) are shown in red. Tracker-computed feature locations are shown in green.}
\end{figure}

%This file was auto-generated by testing_6_compileResults.py (Author: Bryan Poling)
%on October 30, 2013. Modifying this file directly may be unwise, as these changes
%will be wiped out if the file is re-generated. Include this file with
%\input{Mode1ResultsTable_MeanL1Error.tex} where the table is desired.
\begin{table*}[htb]
\captionsetup{margin=10pt,font=small,labelfont=bf}
\caption{Mean L1 trajectory error after 30 frames of tracking. Lower is better.}
\centering
\scriptsize{
\tabcolsep=0.30cm
%\rowcolors{5}{white}{lightgray}
\begin{tabular}{l l | c c c c c c c c | c |}
\cline{3-11}
& & \multicolumn{8}{|c|}{\textbf{Video Number}} & \multicolumn{1}{|c|}{Average} \\
\cline{3-10}
& & \multicolumn{1}{|c|}{1} & \multicolumn{1}{|c|}{2} & \multicolumn{1}{|c|}{3} & \multicolumn{1}{|c|}{4} & \multicolumn{1}{|c|}{5} & \multicolumn{1}{|c|}{6} & \multicolumn{1}{|c|}{7} & \multicolumn{1}{|c|}{8} & \\
\hline
\multicolumn{1}{|c|}{\multirow{9}{*}{\begin{sideways}\textbf{Tracker}\end{sideways}}}
                       & \multicolumn{1}{|l|}{KLT}                                   &         959.6   &         2484.0  &         958.3   &         1242.4  &         1630.2  &         1391.4  &         2105.0  &         4387.6  &         1894.8  \\ \cline{2-11}
\multicolumn{1}{|c|}{} & \multicolumn{1}{|l|}{1st-Order Descent}                     & \textbf{92.5  } &         137.8   &         159.5   &         273.2   &         87.5    &         198.4   &         70.6    &         685.7   &         213.2   \\ \cline{2-11}
\multicolumn{1}{|c|}{} & \multicolumn{1}{|l|}{LDOF}                                  &         508.0   &         408.6   &         898.5   &         385.2   &         104.9   & \textbf{122.3 } &         256.1   &         721.3   &         425.6   \\ \cline{2-11}
\multicolumn{1}{|c|}{} & \multicolumn{1}{|l|}{Multi-Tracker - Emp Dim - Uncentered}  &         104.1   &         139.3   &         128.8   &         241.9   &         75.2    &         136.9   &         58.8    &         305.5   &         148.8   \\ \cline{2-11}
\multicolumn{1}{|c|}{} & \multicolumn{1}{|l|}{Multi-Tracker - Emp Dim - Centered}    &         102.9   & \textbf{115.3 } & \textbf{108.3 } & \textbf{226.8 } & \textbf{69.7  } &         128.6   & \textbf{54.1  } & \textbf{292.5 } & \textbf{137.3 } \\ \cline{2-11}
\multicolumn{1}{|c|}{} & \multicolumn{1}{|l|}{Multi-Tracker - Nuc Norm - Uncentered} &         106.8   &         134.0   &         131.7   &         243.9   &         73.4    &         132.6   &         58.4    &         293.9   &         146.8   \\ \cline{2-11}
\multicolumn{1}{|c|}{} & \multicolumn{1}{|l|}{Multi-Tracker - Nuc Norm - Centered}   &         103.5   &         137.7   &         141.5   &         243.9   &         73.4    &         135.5   &         60.3    &         341.0   &         154.6   \\ \cline{2-11}
\multicolumn{1}{|c|}{} & \multicolumn{1}{|l|}{Multi-Tracker - Exp Fact - Uncentered} &         103.4   &         169.7   &         131.3   &         246.1   &         74.6    &         134.3   &         62.5    &         307.3   &         153.7   \\ \cline{2-11}
\multicolumn{1}{|c|}{} & \multicolumn{1}{|l|}{Multi-Tracker - Exp Fact - Centered}   &         102.9   &         167.0   &         129.1   &         245.0   &         73.2    &         133.4   &         58.9    &         302.5   &         151.5   \\ \hline
\end{tabular}
}
\label{table:tracker mode1 results - Mean L1 Error}
\end{table*}
  %Label: table:tracker mode1 results - Mean L1 Error
%\input{Table_L2_Errors.tex}  %Label: tab:L2 Errors
%This file was auto-generated by testing_6_compileResults.py (Author: Bryan Poling)
%on October 30, 2013. Modifying this file directly may be unwise, as these changes
%will be wiped out if the file is re-generated. Include this file with
%\input{Mode2ResultsTable.tex} where the table is desired.
\begin{table*}[htb]
\captionsetup{margin=10pt,font=small,labelfont=bf}
\caption{Average number of frames between feature re-initializations. Higher is better.}
\centering
\scriptsize{
\tabcolsep=0.345cm
%\rowcolors{5}{white}{lightgray}
\begin{tabular}{l l | c c c c c c c c | c |}
\cline{3-11}
& & \multicolumn{8}{|c|}{\textbf{Video Number}} & \multicolumn{1}{|c|}{Average} \\
\cline{3-10}
& & \multicolumn{1}{|c|}{1} & \multicolumn{1}{|c|}{2} & \multicolumn{1}{|c|}{3} & \multicolumn{1}{|c|}{4} & \multicolumn{1}{|c|}{5} & \multicolumn{1}{|c|}{6} & \multicolumn{1}{|c|}{7} & \multicolumn{1}{|c|}{8} & \\
\hline
\multicolumn{1}{|c|}{\multirow{9}{*}{\begin{sideways}\textbf{Tracker}\end{sideways}}}
                       & \multicolumn{1}{|l|}{KLT}                                   &         10.6    &         7.8     &         13.5    &         9.6     &         7.6     &         9.5     &         7.8     &         2.0     &         8.6     \\ \cline{2-11}
\multicolumn{1}{|c|}{} & \multicolumn{1}{|l|}{1st-Order Descent}                     &         38.9    &         30.3    &         68.7    &         27.7    &         34.0    &         41.1    &         44.1    &         3.9     &         36.1    \\ \cline{2-11}
\multicolumn{1}{|c|}{} & \multicolumn{1}{|l|}{LDOF}                                  &         8.8     &         12.0    &         13.7    &         20.4    &         25.5    &         68.4    &         23.1    &         6.7     &         22.3    \\ \cline{2-11}
\multicolumn{1}{|c|}{} & \multicolumn{1}{|l|}{Multi-Tracker - Emp Dim - Uncentered}  &         73.4    &         35.3    &         111.5   &         55.6    & \textbf{70.2  } &         102.3   &         63.7    &         14.0    &         65.8    \\ \cline{2-11}
\multicolumn{1}{|c|}{} & \multicolumn{1}{|l|}{Multi-Tracker - Emp Dim - Centered}    &         74.3    & \textbf{38.2  } &         111.5   & \textbf{58.5  } &         69.1    &         104.4   &         65.6    &         14.0    &         67.0    \\ \cline{2-11}
\multicolumn{1}{|c|}{} & \multicolumn{1}{|l|}{Multi-Tracker - Nuc Norm - Uncentered} &         77.2    &         35.3    &         108.4   &         53.8    &         62.1    &         105.5   & \textbf{68.5  } &         13.6    &         65.6    \\ \cline{2-11}
\multicolumn{1}{|c|}{} & \multicolumn{1}{|l|}{Multi-Tracker - Nuc Norm - Centered}   & \textbf{78.2  } &         33.8    & \textbf{114.9 } &         54.3    &         66.9    & \textbf{109.0 } &         65.6    &         13.8    & \textbf{67.1  } \\ \cline{2-11}
\multicolumn{1}{|c|}{} & \multicolumn{1}{|l|}{Multi-Tracker - Exp Fact - Uncentered} &         76.2    &         34.3    & \textbf{114.9 } &         54.3    &         66.9    &         98.3    & \textbf{68.5  } &         13.6    &         65.9    \\ \cline{2-11}
\multicolumn{1}{|c|}{} & \multicolumn{1}{|l|}{Multi-Tracker - Exp Fact - Centered}   &         74.3    &         34.8    &         111.5   &         53.0    &         68.0    & \textbf{109.0 } &         65.6    & \textbf{14.3  } &         66.3    \\ \hline
\end{tabular}
}
\label{table:tracker mode2 results}
\end{table*}
  %Label: table:tracker mode2 results

\subsection{Qualitative Experiments on Real Videos}
\label{sec:Qualitative experiments on real videos}
In our tests on real video sequences containing low-quality features, single-feature tracking does not provide acceptable results. When following a non-distinctive feature, the single-feature energy function often flattens out in one or more directions. A tracker may move in any ambiguous direction without realizing a better or worse match with the features template. This results in the tracked location drifting away from a features true location (i.e. ``wandering''). This is not a technical limitation of one particular tracking implementation. Rather, it is a fundamental problem due to the fact that the local imagery in a small neighborhood of a feature does not always contain enough information to deduce the features motion between frames. This claim can be verified by attempting to hand-register low-quality features by only looking at a small neighborhood of the features last known location.

In these situations, our method infers the global motion of the scene from the observable features and uses it to assist in locating the low-quality features. This yields better overall tracking results in hard-to-track videos. Fig.~\ref{fig:Real Video} shows characteristic results from our tests. Several real videos are included in the supplemental material with results from the OpenCV Lucas Kanade tracker, and from our method.

\subsection{Experiments on Synthetically Degraded Videos}
\label{sec:Quantitative experiments on videos with added noise}
For this experiment, we collected 8 video sequences of variable length in favorable lighting conditions. We used a Lucas Kanade tracker to track many features and manually verified and corrected the individual trajectories. Features come and go in these sequences (we do not assume all features persist through the entire sequence). These videos include $6$ rigid environments as well as one video with multiple rigid bodies (video $7$) and one video with a deformable body (video $8$). The sequences range in length from $97$ frames to $289$ frames. On average, they are $210$ frames each and contain over $6000$ feature-frames each (this is the sum of each tracked features lifespan, measured in frames).

We degraded each video sequence by first darkening and adding noise to each frame, followed by applying a strong Gaussian blur to each frame. After this we added additional Gaussian noise. Adding noise before and after blurring gave the effect of noise at different scales (harder to deal with than per-pixel noise only). The test videos are included in the supplementary material.

For our comparison, we ran each tracker in two different modes. In the first mode we initialized each feature with its ground-truth location and re-initialized features when they wandered more than $10$ pixels from ground truth. We recorded the average number of frames between feature re-initializations. In the second mode, we only tracked the features that were visible in frame 0, and features were never re-initialized. We looked at the mean $L_1$ difference between the output trajectories and ground truth after $30$ frames.

As a reference, we compared against the pyramidal Lucas Kanade tracker in the current OpenCV release (2.4.3). For a more recent comparison, we used LDOF (Large Displacement Optical Flow \cite{LDOF_2011}) to generate dense flow fields for each sequence and we interpolated these flow fields to generate long-run trajectories for features. We also implemented our own single-feature gradient descent tracker (with an absolute value loss function). We present results for our rank-constrained tracker with the three previously introduced penalty functions. For each penalty function we present results with and without centering the history matrix $\bM$. In this experiment, whenever our algorithm is run with the penalty term, we use a weak constraint. All trackers were run on grayscale video. The results of this experiment are presented in Tables \ref{table:tracker mode1 results - Mean L1 Error} and \ref{table:tracker mode2 results}. An additional set of tests (on shorter video sequences) is included in the supplementary material.

\subsection{Analysis of Results}
\label{sec:Analysis of Results}
From Tables \ref{table:tracker mode1 results - Mean L1 Error} and \ref{table:tracker mode2 results}, we can see that imposing our weak rank constraint significantly improves overall tracking ability, with all three rank regularizers that we tested showing improved tracking performance. Comparing the Lucas Kanade results to the results of our single-feature gradient descent tracker, we see a very large gap in performance. The core differences between these two algorithms are the definitions of $\psi$ (squared error vs. absolute value) and the method of optimization employed. This performance difference supports our previous claim that the $2^{\text{nd}}$-order optimization technique used to accelerate convergence in the Lucas Kanade algorithm can be unreliable when tracking poor-quality features.

\section{Conclusion}
\label{sec:Conclusion}
Rank constraints have been successfully applied to several problems in computer vision, including motion segmentation and optical flow estimation. We have expanded on this previous work by developing a feature tracking framework which allows these constraints to be reliably used to assist in the tracking of features in rigid environments as well as in more general, non-rigid settings. The framework we presented permits these constraints to be imposed forcefully, allowing one to track features on a rigid object even if some features are occluded, or weakly, where the constraints are only used to help locate poor-quality features that cannot be tracked on their own. We showed that the weak constraint can yield significant gains in tracking performance, even in non-rigid scenes (with multiple or deformable objects) The framework we presented is completely causal and does not require explicitly modeling structure or motion in a scene. Furthermore, the algorithm we proposed is not prohibitively computationally expensive (real-time performance has been achieved). Our results provide evidence that when tracking features in low-quality video (especially in a rigid or semi-rigid scene), a $1^{\text{st}}$-order descent scheme is more robust than $2^{\text{nd}}$ order methods used in standard Lucas-Kanade trackers, and applying rank regularizers to track a cohort of features results in better performance than classical single-feature tracking.

%We presented results showing the dominance of our method over classical singe-feature tracking in scenes containing low-quality features. We also showed that our formulation compares favorably against some alternative implementations of the rigid-body dimensionality constraints.

{\small
\bibliographystyle{ieee}
\bibliography{refs_10_21_13_conf}
}

\end{document}